\documentclass{article}

\PassOptionsToPackage{numbers, compress}{natbib}

 \usepackage[dblblindworkshop, final]{neurips_2025}

\usepackage{wrapfig}
\usepackage[utf8]{inputenc} 
\usepackage[T1]{fontenc}    
\usepackage{hyperref}       
\usepackage{url}            
\usepackage{booktabs}       
\usepackage{amsfonts}       
\usepackage{nicefrac}       
\usepackage{microtype}      
\usepackage{xcolor}         
\usepackage{amsmath}
\usepackage{graphicx}
\usepackage{acronym}
\usepackage{array}
\title{MC-Risk: Multi-Component Risk Fields for Risk Identification and Motion Planning}

\author{%
   Maximilian Link \thanks{Equal contribution.}\\
  Technical University of Munich \\
  Munich, Germany\\
  \texttt{link.maximilian@tum.de} \\
  \And
  Yingjie Xu*  \\
  Technical University of Munich \\
  Munich, Germany\\
  \texttt{yingjie.xu@tum.de} \\
  \And
  Yingbai Hu \\
  The Chinese University of Hong Kong \\
  Hong Kong, China \\
  \texttt{yingbai.hu@tum.de} \\
  \And
  Yinlong Liu \thanks{Corresponding author.}\\
  City University of Macau \\
  Macau, China \\
  \texttt{ylliu@cityu.edu.mo} \\
}

\begin{document}

\maketitle

\begin{abstract}
We present \textbf{MC-Risk}, a planner-aligned, multi-component risk field on a bird’s-eye-view grid that yields early, calibrated, and class-aware risk localization. MC-Risk linearly composes three interpretable modules: (i) a motorized-agent field that fuses a black-box multimodal trajectory predictor with an analytic Gaussian-torus construction whose lateral width grows with speed/curvature and whose height attenuates with look-ahead; (ii) a VRU risk field that replaces isotropic pedestrian blobs with a forward-biased anisotropic kernel aligned to heading and speed; and (iii) a road penalty field that exploits full HD-map topology, imposing an off-road penalty and lane-aware risk exposure for same/opposite directions. We conduct, to our knowledge, the first standardized quantitative evaluation of a risk-field formulation on RiskBench’s collision subset. MC-Risk attains the best overall risk localization and the earliest hazard indication. Finally, we demonstrate a plug-and-play planning interface by using the field as an MPC cost density, enabling risk-aware trajectory generation without additional training.
\end{abstract}    
\section{Introduction}

Building autonomous vehicles that can localize risk early and reliably in cluttered traffic is a core safety challenge. Most planners need explainable, stable risk/cost evaluation with high spatial resolution that reflects different traffic agents (cars, trucks, motorcycles, pedestrians) and rich map semantics. A wide range of approaches has emerged: classical potential-risk-field formulations provide spatially planner-compatible abstractions of scenario danger~\cite{nister2019safety, wang2016driving, wang2015driving, liu2024research}; data-driven risk assessment methods estimate collision likelihoods or required decelerations from perception and forecasting pipelines~\cite{wang2025risk, li2024spatiotemporal, SHI2022106836, HU2021115041}, often leveraging efficient transformer backbones to improve robustness and latency
\cite{10292927}; and recent works couple multimodal forecasting with risk fields to propagate behavioral uncertainty over space and time \cite{jiang2024edrf}. Equally important, standardized benchmarks are needed to measure both the quality and timeliness of risk perception and identification. RiskBench addresses this by providing a scaled dataset and a suite of risk-centric metrics~\cite{kung2024riskbench}.

Despite the progress, three points limit the current approaches. First, the balance between interpretability and scenario coverage: learning-based risk predictors expose comprehensive but black-box results, while classical field theories often underfit real road diversity by using class-agnostic envelopes (e.g., fixed vehicle tubes or isotropic pedestrian blobs) that neglect velocity, possible future trajectory, and forward intent~\cite{wang2015driving}, so finding the balance between these two aspects is still under exploring. Second, map semantics: penalties for off-road regions and opposing lanes are frequently implemented as a fixed buffer, which either over-penalize plausible maneuvers or weaken the ability to identify true risks~\cite{liu2024research}. Third, evaluation rigor: to the best of our knowledge, no prior risk-field formulation has been quantitatively evaluated on a standardized and public benchmark. Most reports are qualitative, lacking actor-set alignment~\cite{jiang2024edrf}. Consequently, it remains difficult to obtain risk maps that are early, comprehensive, calibrated, and to demonstrate these properties under a standardized benchmark.

We introduce \textbf{MC-Risk}, a multi-component risk-field representation on a BEV grid that linearly composes three interpretable modules:
(i) a motorized-agent field (MAF) that combines a black-box multimodal trajectory predictor with an explainable Gaussian-torus construction. Its lateral width varies with speed and curvature, while height attenuates with look-ahead distance, yielding earlier and more precise risk identification.
(ii) a Vulnerable Road User risk field (VRF) that replaces symmetric blobs with a forward-biased, anisotropic kernel aligned to heading and speed, improving risk identification recall for approaching pedestrians or other VRUs.
(iii) a road penalty field (RPF) that leverages full HD-map topology, including off-road, same-direction adjacent lanes, and opposite-direction lanes, to encode a structure-aware risk field rather than manually-defined buffers.

We integrate MC-Risk into RiskBench~\cite{kung2024riskbench} and conduct standardized evaluations on its collision subset, where risk is most prominent. We compute actor-level risk by taking the maximum field value over each actor and thresholding. Across baselines from four families (rule-based, forecast+check, collision anticipation, behavior prediction), MC-Risk achieves the strongest overall risk localization and the earliest hazard indication. Additionally, we integrate the risk map into a standard MPC objective function as a plug-and-play cost, enabling risk-aware motion planning without any additional training.

\paragraph{Contributions of this work.}
\begin{itemize}
    \item 
 We propose a modular, planner-aligned risk-field formulation that couples black-box prediction with explainable analytical fields: velocity-/curvature-conditioned MAF, forward-biased VRF, and topology-aware RPF.  
    \item 
 We conduct a RiskBench-compliant evaluation with a visibility adapter and a unified metric suite (OT-F1, OT-F1-T at 1/2/3\,s, wMOTA, PIC), reported on the collision subset, achieving the best overall performance. 
    \item 
 We propose a planning interface that treats the aggregated field as a cost density for MPC, enabling risk-aware trajectory selection without additional training.
\end{itemize}

\section{Related Works}
\label{sec:background}

\paragraph{Time/Distance Proximity Methods.}
Classical indicators—including time headway (TH) \cite{jansson2005collision}, time-to-collision (TTC) \cite{hayward1972near}, and their temporally aggregated variants TET/TIT \cite{minderhoud2001extended}, PET \cite{allen1978analysis}, etc. offer low-cost signals of closing gaps and exposure \cite{puphal2019probabilistic,wang2021review}. Although intuitive and efficient, they assume simple kinematics, lack uncertainty modeling, and require threshold tuning, limiting the application in multi-agent, multimodal settings.

\paragraph{Formal Safety–Guarantee Methods.}
Probabilistic survival processes fuse predicted position uncertainty into an integrated collision risk rate (Risk-Spot Detector, RSD) \cite{puphal2019probabilistic}. Responsibility-Sensitive Safety (RSS) derives longitudinal/lateral safe distances and scores risk by violation margin \cite{guo2022responsibility}. Hierarchical analytic-network processes quantify scenario risk from layered static/dynamic factors with fuzzy uncertainty handling \cite{wei2024risk}. These methods provide principled safety guarantees but can be conservative or computationally heavy. 

\paragraph{Potential-Field Methods.}
Risk is often encoded as spatial fields that repel the ego from hazards while attracting it to goals. Liu et al.\ define distinct fields for vehicles, pedestrians, and road boundaries, combined into a unified driving-risk field with optional attractive terms for navigation \cite{liu2024research}. NVIDIA’s Safety Force Field (SFF) and its differentiable reproductions compute a safety potential over claimed trajectory sets and apply its negative gradient as an intervention force \cite{nister2019safety,suk2022rationale}. These fields yield a unified representation but can be sensitive to parameterization and not understanding the scenario comprehensively. EDRF \cite{jiang2024edrf} improves realism over purely kinematic fields based on multimodal predicted trajectories (e.g., through QCNet \cite{zhou2023QCNet}, yet its published formulation focuses only on motor vehicles and shows limited map semantics.

\paragraph{Learning-Based Methods.}
Collision-anticipation networks (e.g., DSA \cite{chan2016DSA}, RRL \cite{zeng2017RRL}) score collision likelihood from spatiotemporal features without explicit overlap checks. Trajectory predictors (Social-GAN \cite{gupta2018SocialGAN}, MANTRA \cite{marchetti2020MANTRA}, QCNet \cite{zhou2023QCNet}) enable collision checking by future path overlap. Behavior-influence models (BP/BCP) infer risk from predicted deviations to the ego intention via attention or causal masking \cite{li2020BP,li2020BCP}. Recent systems also embed risk into transformer decoders \cite{wang2025risk} or use LLMs for evaluation and reasoning \cite{you2025comprehensive,zhou2024safedrive}. Complementary advances in perception, such as 
event/frame fusion for detection, offers robust inputs to downstream risk estimation
\cite{9546775,cao2024embracing}.
These methods achieve strong anticipation but are often black-box and may lack planner-ready structure.

We bridge these gaps by coupling a black-box multimodal predictor with an explainable field to produce a class-aware, topology-informed risk map that comprehensively covers all traffic participants and road geometry to remain planner-aligned.

\section{Methods}
\label{sec:methods}
Our goal is to construct a calibrated BEV risk density map $R_{scene}(x,y)$ that motion planners can utilize directly. This section formalizes MC-Risk by composing probability (from trajectory prediction) and consequence (class- and velocity-conditioned severity) into motorized agents, VRU, and road fields. We begin with an overview of the overall definition in Sec.~\ref{sec:Overview}, followed by detailed descriptions of the three key components: the motorized-agent field based on trajectory prediction in ~\ref{sec:MAF}; the forward-biased anisotropic VRU risk field in Sec.~\ref{sec:VRF}; the road penalty field via off-road costs in Sec.~\ref{sec:RPF}.
\subsection{Overview}
\label{sec:Overview}
\begin{figure*}[t!]
    \centering
    \includegraphics[width=\textwidth]{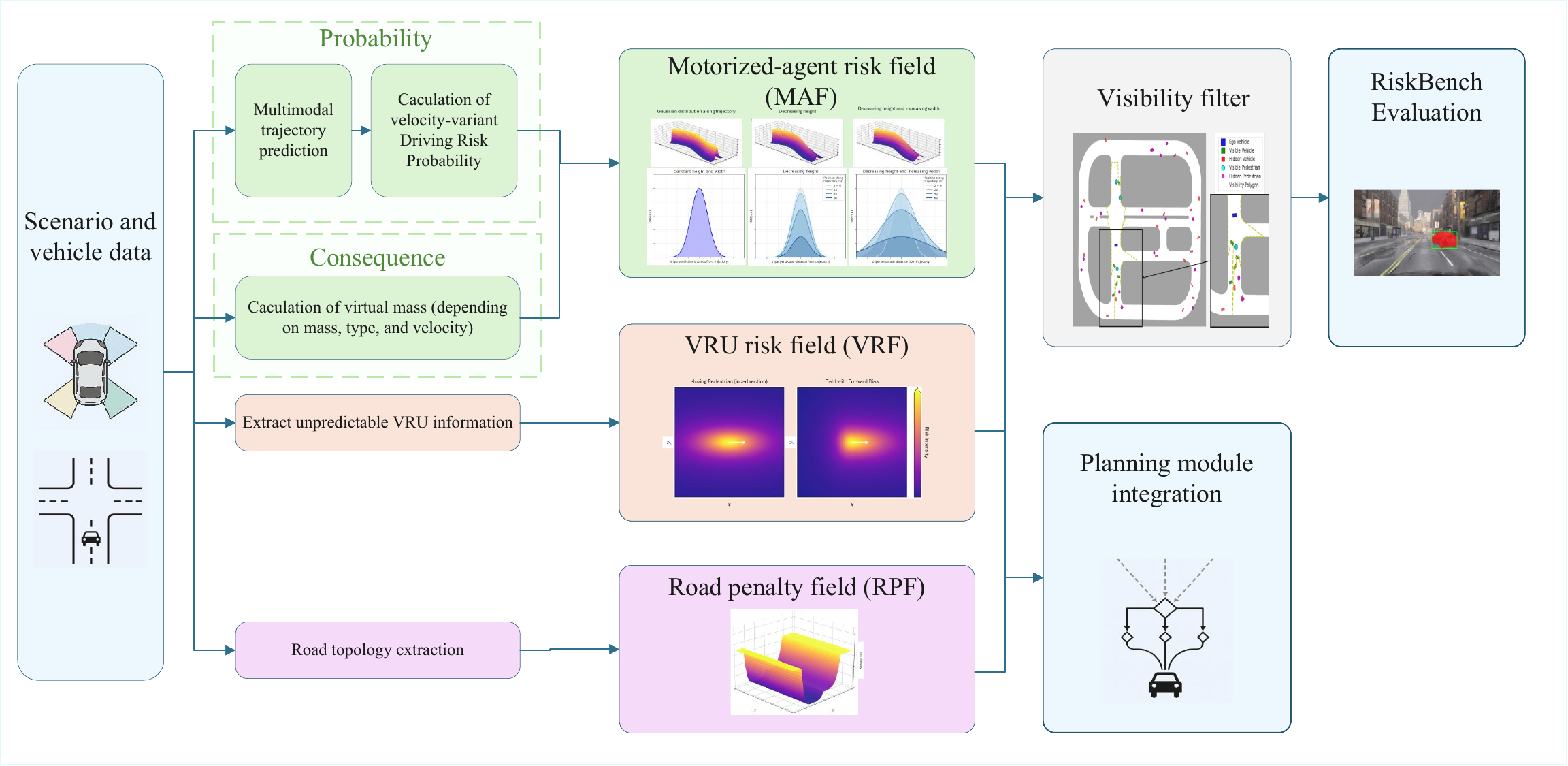}
    \caption[MC-Risk framework]{Overall pipeline of \textbf{MC-Risk}. Scene and vehicle data feed a multimodal trajectory predictor to compute a velocity-variant driving risk probability, which is combined with a virtual-mass consequence term to form the motorized-agent risk field (MAF). In parallel, road topology is extracted to build a topology-aware road penalty field (RPF), and VRU state/heading are used to construct a forward-biased anisotropic VRU field (VRF) since they are relatively unpredictable. The three fields are linearly composed into a BEV risk map, filtered by a visibility adapter, evaluated with RiskBench metrics, and optionally injected as a cost into an MPC planner. }   \label{fig:framework}
\end{figure*}

We discretize a local BEV grid \(\Omega \subset \mathbb{R}^2\) around the ego vehicle. As suggested by Puphal~et al. \cite{Puphal_2019}, for visualization or a conceptual combined use outside of the ego-vehicle risk assessment, the field components can be superimposed, so our MC-Risk comprises component fields:
\begin{equation}
  R_{scene}(x,y) = \sum_{i=1}^{N_{mot}} \mathrm{MAF}_i(x,y) \ + \ \sum_{k=1}^{N_{VRU}} \mathrm{VRF}_k(x,y) \ + \ \mathrm{RPF}(x,y) \ ,\qquad (x,y)\in\Omega,
  \label{eq:combined_field} 
\end{equation}
where $N_{mot}$ is the number of motorized agents and $N_{VRU}$ is the number of vulnerable road users in the scene. We denote the field for vehicle $i$ by $\mathrm{MAF}_i(x,y)$ and the field for pedestrian $k$ by $\mathrm{VRF}_k(x,y)$. Each component is a sum over instances (dynamic agents) or static primitives. An overview of the full pipeline is shown in Fig.~\ref{fig:framework}.

\subsection{Motorized-agent fields (MAF)}
\label{sec:MAF}
\begin{figure*}[t!]
    \centering
    \includegraphics[width=\textwidth]{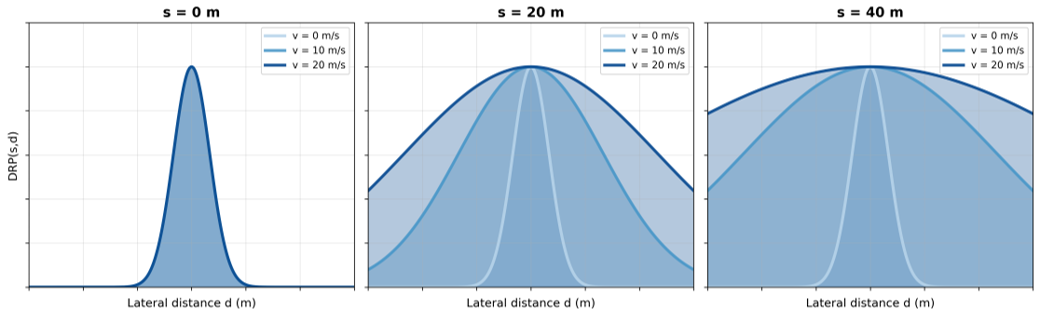}
    \caption[Velocity-variant lateral dispersion]{Velocity-variant width in MAF cross-sections. Lateral slices at three path distances ($s{=}0,20,40$\,m) for speeds $v{=}0,10,20$\,m/s. Higher speeds produce larger $\sigma(s)$ via the $k_v|v(s)|$ term in Eq.~\eqref{eq:velvar}.}
    \vspace{-2.0em}

    \label{fig:velvar}
\end{figure*}

For each motorized agent $i$ (cars, motorcycles, trucks), we couple a black-box multimodal trajectory predictor with an explainable analytical field. Concretely, a learned predictor (e.g., QCNet \cite{zhou2023query} or any equivalent prediction network) outputs a set of $M$ candidate trajectory centerlines $\{\gamma_i^{(m)}\}_{m=1}^M$ with probabilities $\{p_i^{(m)}\}_{m=1}^M$. We then transform these hypotheses into a continuous, planner-aligned risk density via a Gaussian-torus construction in Frenet coordinates, following the Enhanced Driving Risk Field (EDRF) principle \cite{jiang2024edrf} that risk equals \textbf{accident probability} times \textbf{accident consequence} \cite{puphal2019probabilistic}. Gaussian parameterizations are widely used in perception-related tasks for compact geometry encoding,
which we adopt here for risk field design \cite{9990918}.

Let $(s,d)$ denote longitudinal/lateral Frenet coordinates of a BEV location $(x,y)\in\Omega$ relative to $\gamma_i^{(m)}$. The MAF for hypothesis $m$, followed by driving-risk probability (DRP) \cite{kolekar2020human}, is a line integral of Gaussian cross-sections along the path:

\begin{equation}
    \mathrm{MAF}_i^{(m)}(x,y) = \mathrm{MAF}_i^{(m)}(s,d) = a(s) \cdot \exp{\left(-\frac{d^2}{2 {\sigma(s)}^2}\right)}
\label{eq:drp}
\end{equation}

The cross-section are shaped by letting the kernel height decay parabolically and the width grow linearly with path length and average curvature:
\begin{equation}
    a(s) = q\,(s - s_{pt})^2, 
    \qquad
    \sigma(s) = (b + k\,\overline{\kappa_{pt}})\,s + c,
\end{equation}
where $s_{pt}$ is the total predicted path length and $\overline{\kappa_{pt}}$ its average curvature. 
The parameter $q$ controls how rapidly the height attenuates to zero at $s=s_{pt}$, while $b,k,c$ control the growth and initial width of the cross-section, encoding greater dispersion for longer and curvier predictions.

\paragraph{Velocity-variant lateral dispersion.}
To anticipate wider reachable sets for fast or aggressive motion (visualized in Fig.~\ref{fig:velvar}), we augment the width function with a speed-sensitive term:
\begin{equation}
    \sigma(s) = (b + k\,\overline{\kappa_{pt}})\,s \;+\; k_{v}\,|v(s)| \;+\; c,
    \label{eq:velvar}
\end{equation}
where $k_v$ scales the velocity-driven widening. This improves early spatial coverage for high-speed segments (longer braking distances) while keeping interpretability; in practice we bound $\sigma(s)\in[\sigma_{\min},\sigma_{\max}]$ to avoid over-diffusion.

\paragraph{Consequence via virtual mass and MAF fusion.}
Similar to EDRF, we model accident consequence by a virtual-mass term
\begin{equation}
    M_i \;=\; m_v\,T\,\big(\alpha\,v(s=0)^{\beta} + \gamma\big),
    \label{eq:EDRF_M}
\end{equation}
with tunable parameters $\alpha,\beta,\gamma$, vehicle mass $m_v$, vehicle type $T$, and speed $v(s=0)$. 
In their formulation, virtual mass is calculated using the current velocity, which will result in a lack of foresight, failing to anticipate the severity of collisions caused by vehicles' potential acceleration maneuvers in the future. Instead we adopt a path-aggregated severity $\bar M_i^{(m)}$:
\begin{equation}
    \bar M_i^{(m)} \;=\; \frac{1}{s_{pt}}\int_{0}^{s_{pt}} M_i(s)\,ds.
    \label{eq:edrf_full}
\end{equation}
Finally, we sum over multimodal hypotheses to obtain the motorized-agent field:
\begin{equation}
    \mathrm{MAF}_i(x,y) \;=\; \sum_{m=1}^{M} p_i^{(m)}\cdot\bar M_i^{(m)}\cdot\mathrm{MAF}_i^{(m)}(x,y),
\end{equation}
which plugs directly into the scene-level composition of Eq.~\eqref{eq:combined_field}.

\subsection{Vulnerable-road-user risk fields (VRF)}
\label{sec:VRF}
\begin{figure*}[t!]
    \centering
    \includegraphics[width=\textwidth]{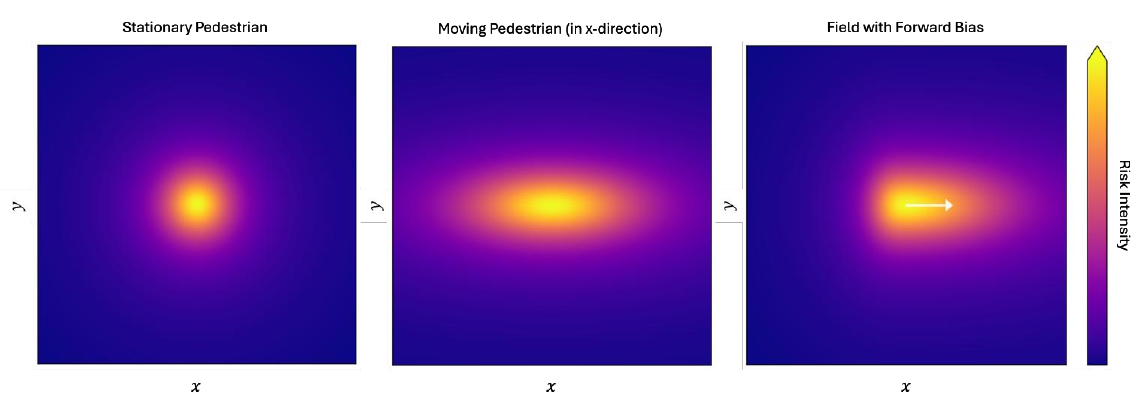}
    \caption[VRF visualization]{VRF visualization. Left: stationary pedestrian—near-isotropic footprint. Middle: moving pedestrian (baseline)—anisotropic ellipse aligned with heading. Right: forward-biased VRF—kernel shifted by $\mu_f$ along heading. Warmer colors indicate higher risk.}
    \vspace{-1.5em}
    \label{fig:vrf_vis}
\end{figure*}
Pedestrians and other VRUs exhibit higher short-horizon unpredictability than motorized agents, so we adopt an anisotropic, speed-adaptive kernel (illustrations in Fig.~\ref{fig:vrf_vis}) inspired by \cite{liu2024research}. Let $\mathbf{p}=(x,y)$, $\mathbf{p_k}$ be the VRU position, $\hat{\mathbf{t}}_k$ its heading unit vector, and decompose the BEV offset as longitudinal offset
$d_{\parallel}=(\mathbf{p}-\mathbf{p_k})\cdot\hat{\mathbf{t}}_k$ and lateral offset
$d_{\perp}=\big\|(\mathbf{p}-\mathbf{p_k})-d_{\parallel}\hat{\mathbf{t}}_k\big\|$.
The baseline elliptical field is
\begin{equation}
    \mathrm{VRF}_k(x,y)=
    \frac{H}{
    \left(\dfrac{d_{\parallel}}{\gamma + k_{pl}\,|v_{\parallel}|}\right)^2
    +
    \left(\dfrac{d_{\perp}}{\delta + k_{pw}\,|v_{\perp}|}\right)^2
    + 1},
    \label{eq:PRF_baseline}
\end{equation}
where $v_{\parallel},v_{\perp}$ are the longitudinal/lateral speed components, $H$ scales overall strength, and $\gamma>\delta$ captures the forward/backward vs.\ sideways compactness.

\paragraph{Forward-bias} To reflect higher risk ahead of a moving VRU, we shift the kernel’s center by a speed-proportional parameter $\mu_f=\lambda_f\,|v_{\parallel}|$ along $\hat{\mathbf{t}}_k$:
\begin{equation}
    \tilde d_{\parallel}=\big(\mathbf{p}-\mathbf{p_k}-\mu_f\hat{\mathbf{t}}_k\big)\!\cdot\!\hat{\mathbf{t}}_k,\quad
    \tilde d_{\perp}=\Big\|\big(\mathbf{p}-\mathbf{p_k}-\mu_f\hat{\mathbf{t}}_k\big)-\tilde d_{\parallel}\hat{\mathbf{t}}_k\Big\|,
\end{equation}

\begin{wrapfigure}{r}{0.5\textwidth}
    \centering
    \includegraphics[width=0.4\textwidth]{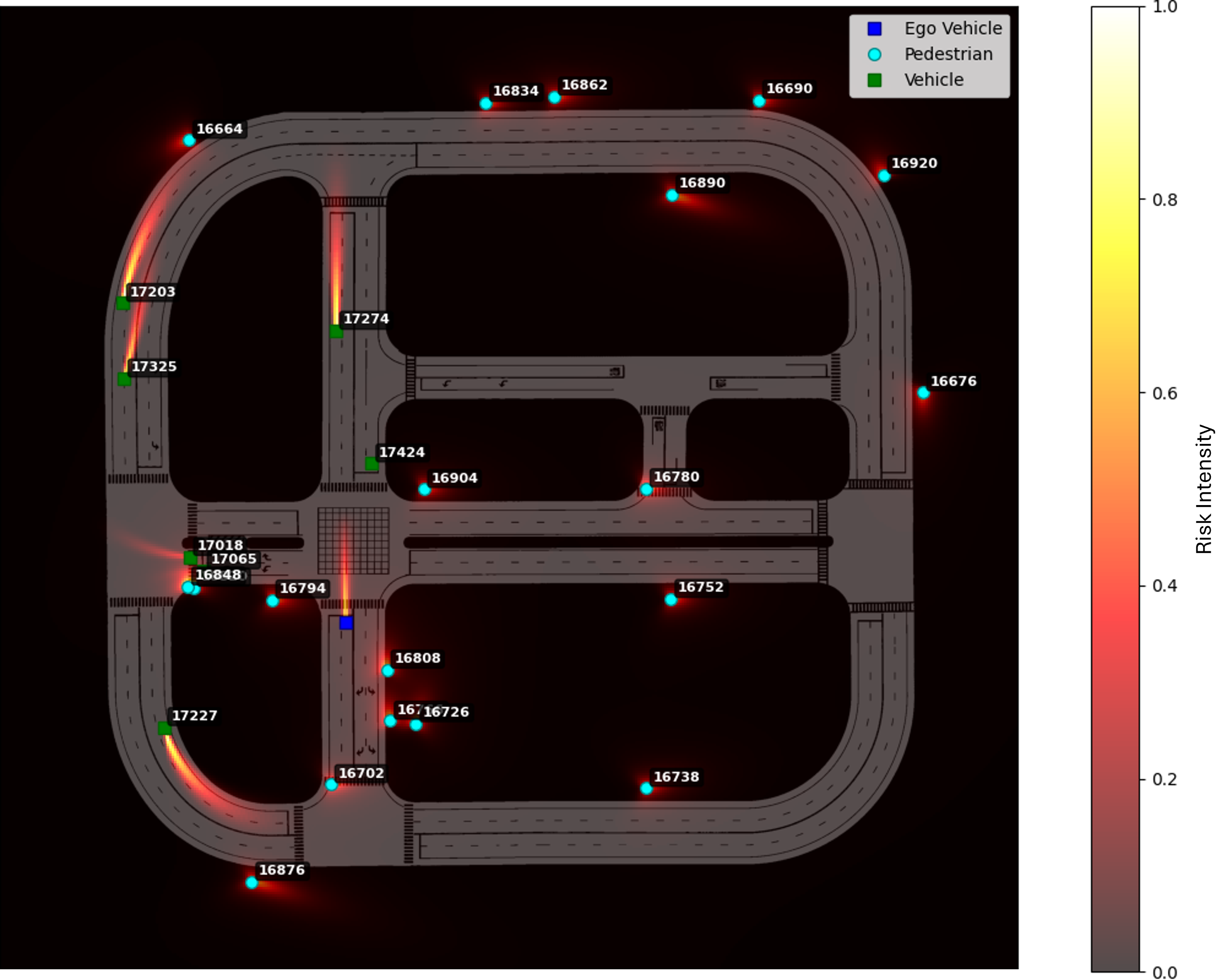}
    \caption{Scene overlay of MAF (vehicles) and VRF (VRUs) with respective ID labeled.}
    \vspace{-0.5em}

    \label{fig:mafvis}
\end{wrapfigure}

This forward-biased formulation increases the ability to detect risk from approaching VRUs by assuming that most VRUs will maintain their current state of motion, and it integrates into Eq.~\eqref{eq:combined_field} alongside $\mathrm{MAF}_i$ and $\mathrm{RPF}$. A scene overlay including both MAF and VRF is shown in Fig.~\ref{fig:mafvis}.

\subsection{Road penalty field (RPF)}
\label{sec:RPF}

Following Liu et al., we encode a topology-aware road penalty independent of dynamic agents; the map partition and resulting field are shown in Fig.~\ref{fig:rpfvis}. Unlike their formulation, which only poses risk based on road boundaries, we use full HD-map topology to include (i) off-road penalties, (ii) same-direction adjacent lanes, (iii) opposite-direction lanes. 
Supposing the road is in $x$ direction, we let $d_{\pm}(y)$ be the signed distance to road boundaries (negative inside). $\mathcal{L}$ is the set of lane centerlines with direction attribute, and we partition $\mathcal{L}$ around the ego’s current lane into $\mathcal{L}_{\text{same}}$ and $\mathcal{L}_{\text{opp}}$. 
Using a single compact expression, the RPF is
\begingroup\setlength{\abovedisplayskip}{6pt}\setlength{\belowdisplayskip}{6pt}

\begin{equation}
\label{eq:RPF_full}
RPF(y) \;=\;
\lambda_{\text{off}}\,\mathbf{1}\!\left[d_{\pm}(y)>0\right]
\;+\!\!\!\!\sum_{\ell\in\mathcal{L}_{\text{same}}}\!\!\!\!
\lambda_{\text{same}}\exp\!\Big(-\tfrac{\mathrm{dist}(y,\ell)^2}{2\sigma_{\text{same}}^2}\Big)
\;+\!\!\!\!\sum_{\ell\in\mathcal{L}_{\text{opp}}}\!\!\!\!
\lambda_{\text{opp}}\exp\!\Big(-\tfrac{\mathrm{dist}(y,\ell)^2}{2\sigma_{\text{opp}}^2}\Big)
\end{equation}
\begin{wrapfigure}{r}{0.5\textwidth}
    \centering
    \vspace{-0.5em}
    \includegraphics[width=0.5\textwidth]{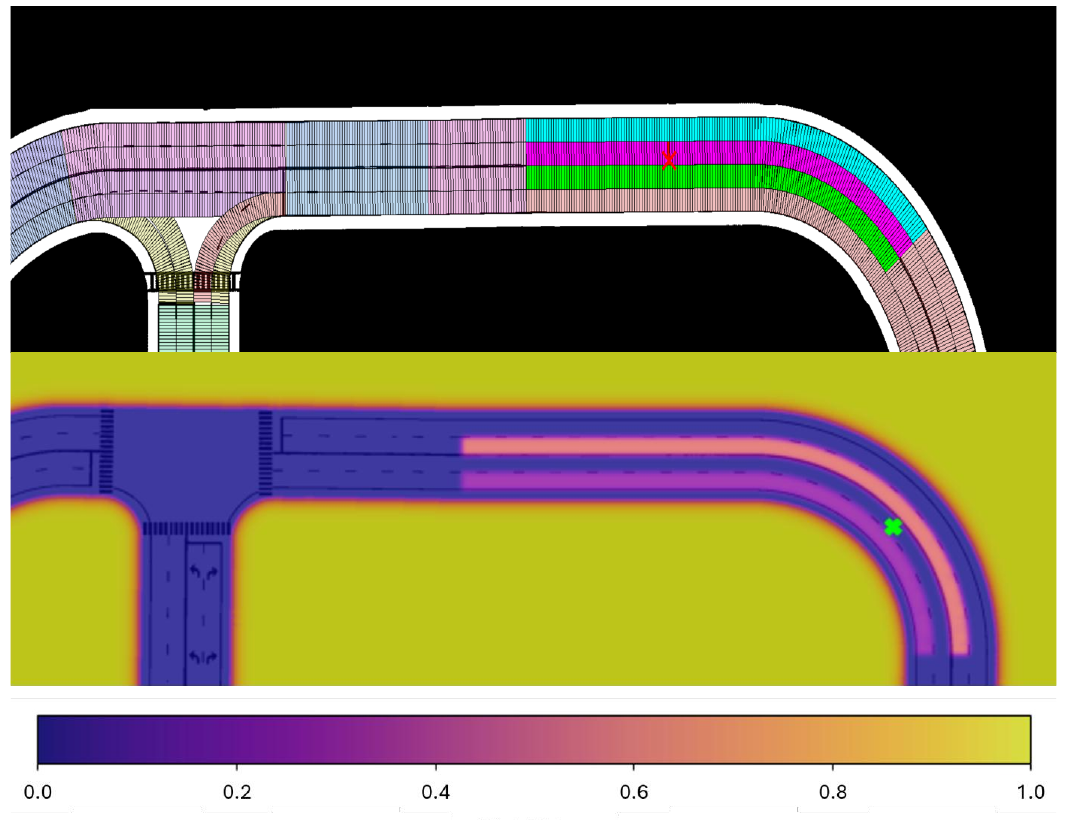}
    \caption[Topology-aware road partition and RPF]{Topology-aware road partition and RPF. Top: CARLA HD-map topology around ego, including the intersection, partitioned into same-/opposite-direction lanes. Bottom: resulting Road Penalty Field (RPF) centered at the ego.}
    \vspace{-2.5em}

    \label{fig:rpfvis}
\end{wrapfigure}
where $\mathbf{1}[\cdot]$ is the indicator function and $\mathrm{dist}(y,\ell)$ is the lateral distance to lane centerline $\ell$. 
Hyperparameters $(\lambda_{\text{off}},\lambda_{\text{same}},\lambda_{\text{opp}},\sigma_{\text{same}},\sigma_{\text{opp}})$ control baseline risk value and spread, typically $\lambda_{\text{off}}>>\lambda_{\text{opp}}>\lambda_{\text{same}}$. The off-road term assigns a large, fixed penalty outside $\mathcal{M}_{\text{road}}$, discouraging driving beyond the drivable boundary. 
The implementation is also compatible in intersections as shown in Fig.~\ref{fig:rpfvis}, and we finally add $\mathrm{RPF}$ to the scene risk field in Eq.~\eqref{eq:combined_field}.

\section{Experiments}
\label{sec:experiment}

\subsection{Experimental settings}
\paragraph{Dataset}
We evaluate on RiskBench, a CARLA-based benchmark by Kung et al. \cite{kung2024riskbench} that couples rich static infrastructure with dynamic agents across diverse scenarios: cut-in, braking, merging, yielding 6916 scenarios with spatiotemporal risk annotations. Despite active research on risk fields and threat assessment, few prior works report standardized, visibility-aware evaluations. Results are often qualitative and omit actor-set alignment. A key contribution of this paper is to adopt RiskBench for per-actor risk quantification, and evaluate the final results with an extended metric suite from RiskBench (\textbf{OT-F1}, \textbf{OT-F1-T} at 1/2/3\,s, \textbf{wMOTA}, \textbf{PIC}). Because risk is most prominent when hazardous events culminate in contact, we restrict evaluation to the collision subset of RiskBench. 

\paragraph{Evaluation metrics}
Following RiskBench \cite{kung2024riskbench} and its adaptations \cite{pao2024potential}, we use actor-aligned metrics for spatial accuracy and temporal consistency. At each frame \(t\), we compute an actor-level risk score \(r_{i,t}=\max_{(x,y)\in\Omega} R_{scene}(x,y)\) as the maximum risk over the BEV grid $\Omega$, and an actor is labeled risky if \(r_{i,t}>\tau\). Sweeping \(\tau\) yields precision \(=TP/(TP{+}FP)\), recall \(=TP/(TP{+}FN)\), and \(F1 = \frac{2 \cdot \text{precision} \cdot \text{recall}}{\text{precision} + \text{recall}}
\). We report \(\mathrm{\textbf{OT\text{-}F1}}=\max_{\tau}F1(\tau)\) and \(\mathrm{\textbf{OT\text{-}F1\text{-}T}}\) is computed over the last \(T\) seconds before the ego’s closest approach to maximum risk to evaluate temporal consistency. The proactivity of risk anticipation is summarized by \(PIC = - \sum_{t=1}^{T} \exp\!\left(-\frac{T-t}{T}\right) \cdot \log\!\big(F_{1,t}\big)\), penalizing a false identification more when ego vehicle is closer to the risk than far from the risk. Robust tracking quality is assessed via \(\mathrm{\textbf{wMOTA}}=1-\frac{\sum_t\!\big[w_p(FN_t{+}IDsw_t^p)+w_n(FP_t{+}IDsw_t^n)\big]}{w_p\,GT_t^p + w_n\,GT_t^n}\), which addresses the imbalance between risky and non-risky samples.

\paragraph{Visibility adapter}

To enable the occlusion-aware and visibility-scoped risk identification, we apply a lightweight line-of-sight adapter used only at evaluation time (Fig.~\ref{fig:occlusionarea}). 
\begin{wrapfigure}{r}{0.5\textwidth}
    \centering
    \includegraphics[width=0.5\textwidth]{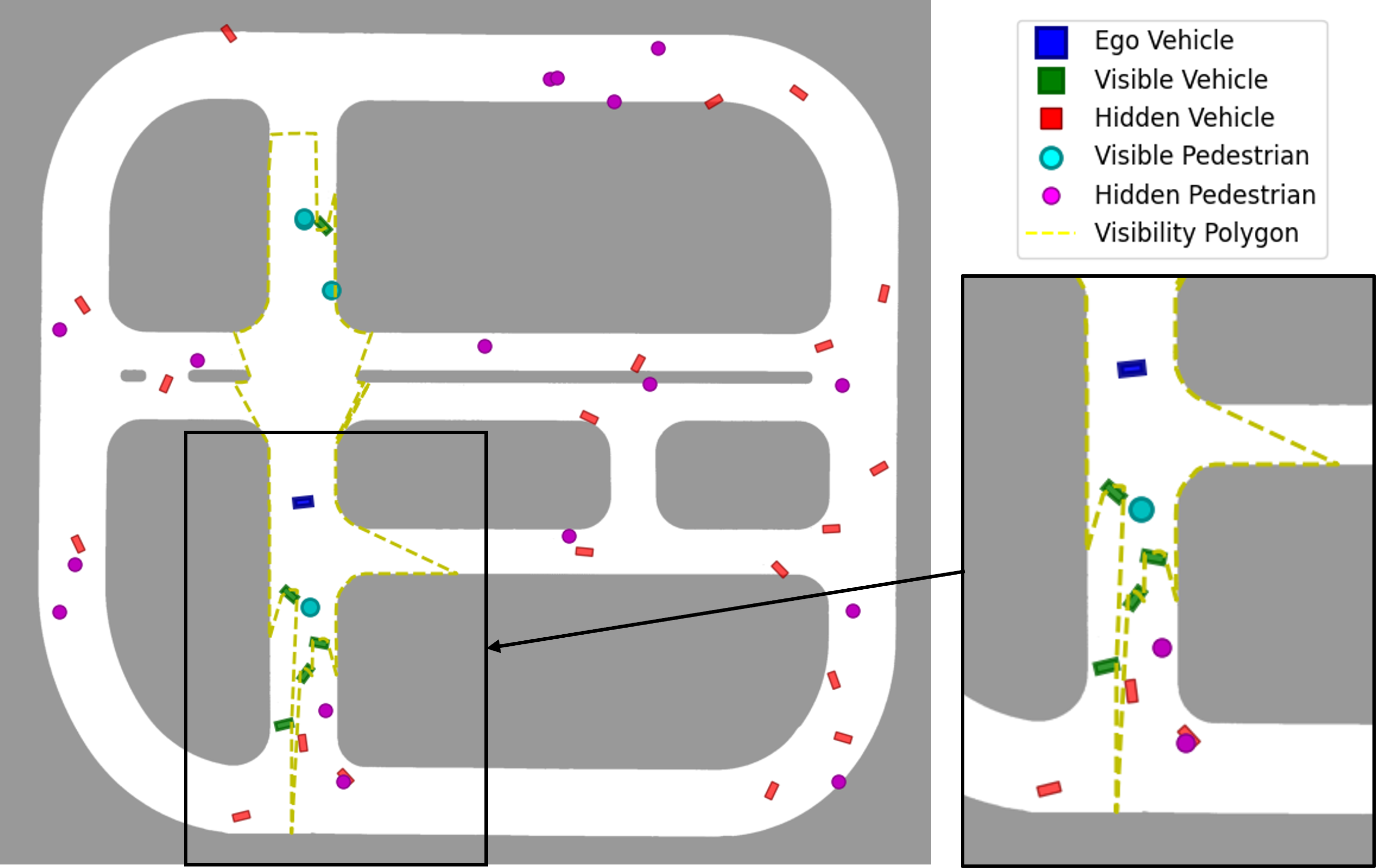}%
    \caption{Visibility filtering utilizing the visibility polygon.}%
    \label{fig:occlusionarea}%
\end{wrapfigure}We create a road mask from HD map data and take non-drivable areas as opaque. Motorized agents are added as dynamic occluders (pedestrians are non-occluding in our setup). From the ego position, we cast a certain number of rays uniformly over \(2\pi\) angle and stop each ray at the first opaque pixel or a fixed max range. The endpoints define a convex visibility polygon. At each time step, we filter both the scene risk map and the actor set by this polygon: actors outside are ignored for metric computation, thus preventing access to off-screen actors, decreasing the false positive rate, and increasing computational efficiency. This adapter aligns our evaluation with a risk model's visibility property while leaving the underlying risk model unchanged.

\subsection{Quantitative results}

\begin{table}[htbp]
\centering
\caption[\textbf{MC-Risk} against RiskBench baselines]{Comparison of  Risk Object Identification (ROI) metrics between \textbf{MC-Risk} and RiskBench baseline models. Baselines are ordered by \textbf{OT-F1} score. The best performer of each metric is marked in bold, and the second best performer is marked in italic.}
\resizebox{\textwidth}{!}{
\begin{tabular}{l!{\vline}c!{\vline}c!{\vline}c!{\vline}c!{\vline}c!{\vline}c}
\toprule
\textbf{Model} & \textbf{OT-F1} ($\uparrow$) & \textbf{wMOTA} ($\uparrow$) & \textbf{PIC} ($\downarrow$) & \textbf{OT-F1-1s} ($\uparrow$) & \textbf{OT-F1-2s} ($\uparrow$) & \textbf{OT-F1-3s} ($\uparrow$) \\
\midrule

Range                   & \textit{65.67\%} & \textbf{77.09\%} & 30.15 & \textit{81.97\%} & 71.79\% & \textit{67.07\%} \\

Kalman filter           & 58.98\% & 72.33\% & \textit{28.49} & 80.33\% & \textit{69.39\%} & 64.25\% \\

DSA                     & 58.19\% & 66.85\% & 39.10 & 79.24\% & 62.78\% & 55.31\% \\

Social-GAN              & 51.97\% & 68.25\% & 46.36 & 64.29\% & 53.98\% & 51.55\% \\

QCNet                   & 51.82\% & 68.35\% & 48.05 & 65.44\% & 56.09\% & 52.26\% \\

MANTRA                  & 51.07\% & 67.75\% & 46.26 & 63.50\% & 53.36\% & 50.39\% \\

BCP                     & 32.56\% & 57.80\% & 64.54 & 38.70\% & 29.93\% & 28.34\% \\

Random                  & 27.75\% & 1.80\%  & 48.40 & 33.55\% & 31.53\% & 30.65\% \\

BP                      & 17.33\% & 53.47\% & 73.10 & 15.69\% & 13.66\% & 14.88\% \\

\midrule
MC-Risk     & \textbf{68.62\%} & \textit{74.03\%} & \textbf{14.78} & \textbf{90.80\%} & \textbf{79.23\%} & \textbf{73.16\%} \\
\bottomrule
\end{tabular}
}
\label{tab:edrf_vs_riskbench_filtered}
\end{table}

We adopt nine reference methods integrated in RiskBench \cite{kung2024riskbench} as baselines, covering four families: (i) \textbf{Rule-based}: \emph{Random} and \emph{Range} (fixed-distance filter) risk object identification provides the simplest baselines. 
(ii) \textbf{Trajectory prediction and collision
 checking}: \emph{Kalman} \cite{thrun2002Kalman}, \emph{Social-GAN} \cite{gupta2018SocialGAN}, \emph{MANTRA} \cite{marchetti2020MANTRA}, and \emph{QCNet} \cite{zhou2023query} first predict agent motion and then simply declare risk via trajectory overlap with ego. 
(iii) \textbf{Collision anticipation}: \emph{DSA} \cite{chan2016DSA} learns to score collision likelihood from spatiotemporal features without explicit overlap tests. 
(iv) \textbf{Behavior prediction-based}: \emph{BP} \cite{li2020BP} and \emph{BCP} \cite{li2020BCP} infer risk from predicted influence on the ego’s intended behavior. 

We use the RiskBench implementations and default settings, apply our visibility adapter, restrict to the collision subset, and evaluate with the unified metric suite described above.

Table~\ref{tab:edrf_vs_riskbench_filtered} shows that \textbf{MC-Risk} delivers the strongest overall risk localization and the earliest hazard indication on the collision subset. It achieves the best \textbf{OT-F1} (68.62\%), surpassing the strongest baseline (Range, 65.67\%) by +2.95 points, while obtaining the lowest \textbf{PIC} (14.78), a {\(\sim\)48\%} reduction versus Kalman filter (28.49), indicating early and reliable risk identification for the ego vehicle to react. Consistency metrics are also the best: \textbf{OT-F1-1s/2s/3s} reach 90.80/79.23/73.16\%, improving upon Range by +8.83/+7.44/+6.09 points, respectively, showing the stability of the model. The only metric where MC-Risk is not the top performer is \textbf{wMOTA}: it ranks second (74.03\%) and is within 3.06 points of Range (77.09\%), suggesting slightly less temporal persistence in tracking robustness but a better precision–recall balance overall. In summary, treating risk identification as a simple downstream of the computed risk field (actorwise max + threshold) already achieves strong performance, validating the MC-Risk field formulation for this subtask.

\subsection{Qualitative results}
\label{sec:qualitative}

\begin{figure*}[t!]%
    \centering%
    \includegraphics[width=\textwidth]{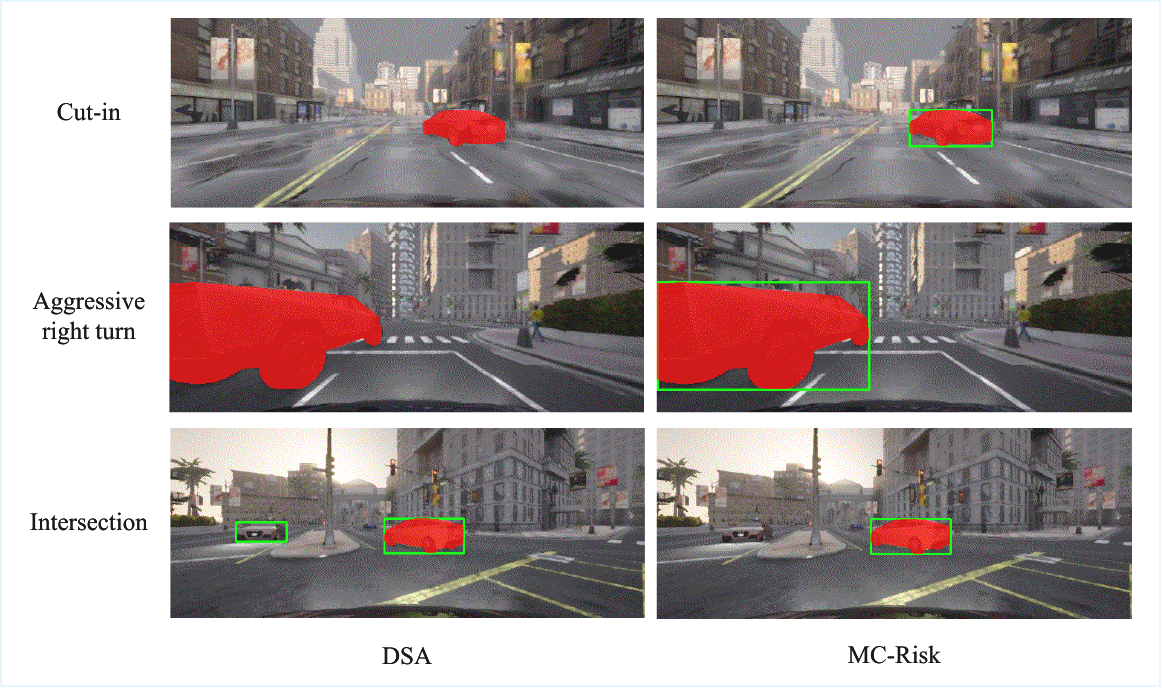}%
    \caption[Qualitative comparison on RiskBench]{Qualitative comparison of risk identification on three representative scenarios. Red overlays denote the ground-truth risky actor from RiskBench, and green boxes mark the predicted risky object at a fixed threshold.}
    \label{fig:qualitative}%
    \vspace{-1.5em}

\end{figure*}%

For qualitative results, we take the DSA model as the baseline. As shown in Figure~\ref{fig:qualitative}, the red layers identify the ground-truth risky object from RiskBench, and the green bounding boxes indicate risky objects derived from our MC-Risk with a tuned threshold. We compare DSA and MC-Risk on three challenging situations: close cut-in, intersection, and aggressive right turn. 

\textbf{Close cut-in:} A neighboring vehicle executes a late merge into the ego lane with small headway and a tight lateral gap. DSA does not identify risk until the intruder is already merging, whereas MC-Risk places early risk along the intruder’s predicted centerline, with velocity-conditioned widening that projects risk exposure into the ego lane ahead of the merge. 

\textbf{Aggressive right turn:} A nearby vehicle starts a sharp right turn across the ego lane at high entry speed and high curvature. DSA again lags and still cannot capture the risk in the displayed time step. MC-Risk expands laterally on high-curvature segments, highlighting the future risk overlap region before the turn is executed. 

\textbf{Intersection:} An oncoming vehicle approaches the intersection, turns right, and decelerates at the same time. The ego vehicle is turning left and trying to enter the same target lane as the other vehicle. Both models can detect the risk, but DSA produces a false positive on the stopping vehicle on the opposite lane; MC-Risk avoid this via reduced risk spread from a low-speed (stopped) agent, yielding a cleaner focus on truly threatening actors. 

These results align with the quantitative gains in early-warning metrics, showing that MC-Risk provides earlier and cleaner localization and avoids false alarms on negative agents.

\subsection{Ablation Studies}

\begin{table}[htbp]
\centering
\caption[MC-Risk component ablations]{Component ablations on \textbf{MC-Risk}. A checkmark indicates the component is enabled. Metrics are still computed on the collision subset with visibility-scoped, actor-aligned evaluation. The best performer of each metric is marked in bold.}
\resizebox{\textwidth}{!}{
\begin{tabular}{l c c c | c c c c c c}
\toprule
\textbf{Model} & \textbf{MAF vel-var} & \textbf{VRF} & \textbf{RPF} & \textbf{OT-F1} ($\uparrow$) & \textbf{wMOTA} ($\uparrow$) & \textbf{PIC} ($\downarrow$) & \textbf{OT-F1-1s} ($\uparrow$) & \textbf{OT-F1-2s} ($\uparrow$) & \textbf{OT-F1-3s} ($\uparrow$) \\
\midrule
MC-Risk (full)                & \checkmark & \checkmark & \checkmark & \textbf{68.62}\% & \textbf{74.03}\% & \textbf{14.78} & \textbf{90.80}\% & \textbf{79.23}\% & \textbf{73.16}\% \\
\midrule
\textit{w/o} vel-variant width (MAF) & $\times$     & \checkmark & \checkmark & 65.09\%  & 72.09\%   & 20.40      & 84.73\%    & 76.05\%        & 69.82\%       \\
\textit{w/o} VRU risk field (VRF)      & \checkmark   & $\times$   & \checkmark & 61.42\%      & 70.81\%      & 22.87    & 71.23\%       & 65.96\%       & 62.02\%       \\
\textit{w/o} road penalty field (RPF)      & \checkmark   & \checkmark & $\times$   & 68.62\%      &  74.01\%      & 19.66    & 90.80\%       & 79.23\%       & 73.16\%       \\
\bottomrule
\end{tabular}}
\label{tab:ablation_mc_risk}
\end{table}

Table~\ref{tab:ablation_mc_risk} quantifies each component’s contribution:

\textbf{MAF velocity-variant width.} Removing the speed-conditioned lateral widening lowers \textbf{OT-F1} from 68.62\% to 65.09\;(-3.53) and \textbf{wMOTA} from 74.03\% to 72.09\;(-1.94), while \textbf{PIC} rises from 14.78 to 20.40\;(+5.62). Lead-time scores also drop, indicating that velocity-variant width is key for predicting risk early along fast segments and reducing the weight of low-speed objects.

\textbf{VRU modeling.} In the VRF ablation, VRUs are also using an MAF to indicate their risks instead of the forward-biased anisotropic VRF. This causes the largest degradation: \textbf{OT-F1} 68.62\% $\rightarrow$ 61.42\;(-7.20), \textbf{wMOTA} 74.03\% $\rightarrow$ 70.81\;(-3.22), and \textbf{PIC} 14.78 $\rightarrow$ 22.87\;(+8.09). Early-warning quality also drops sharply, showing that a forward-biased VRU field is crucial for timely, precise pedestrian/bicycle risk.

\textbf{Road penalty field (RPF).} Removing the static/topology-aware road layer leaves \textbf{OT-F1} (68.62\%) and \textbf{wMOTA} (74.01\%) essentially unchanged, and the lead-time scores are identical to full. However, \textbf{PIC} worsens (14.78 $\rightarrow$ 19.66; +4.88). A plausible explanation is that the RPF layer adds early risk where edge proximity is inherent, which advances correct identifications in time without changing which actor is named or how stable identities are tracked. The effect targets timing more than classification or tracking, hence the integration of RPF shows a clear improvement in the PIC metric.

\subsection{Planning integration}
\label{sec:planning}
In this section, we demonstrate an example to show how the MC-Risk can serve as a planner cost, leaving closed-loop evaluations to future work. 
Let $\mathcal{R}_t(x,y)=R_{\text{scene}}(x,y)$ denote the BEV risk density at time $t$ and let $\xi=\{x_t\}_{t=0}^{H}$ be the ego state trajectory to be planned over horizon $H$ with controls $u=\{u_t\}_{t=0}^{H-1}$. 
We formulate a standard Model Predictive Control (MPC) \cite{kouvaritakis2016model} problem with an additional risk-guidance term:
\begin{equation}
\begin{aligned}
\min_{u,\xi}\;\; 
J(\xi,u) &= \alpha \sum_{t=0}^{H}\hat{\mathcal{R}}_t(x_t)
+\beta \sum_{t=0}^{H-1}\|u_t\|_R^2
+\gamma \sum_{t=1}^{H-1}\|u_t-u_{t-1}\|_S^2 \\
\text{s.t.}\quad
x_{t+1} &= f(x_t,u_t),\\
x_t &\in \mathcal{X},\\
u_t &\in \mathcal{U}.
\end{aligned}
\end{equation}

Here $f$ is a kinematic bicycle model with standard bounds $\mathcal{X},\mathcal{U}$. $R,S$ are positive-definite weights, and $\hat{\mathcal{R}}_t(x_t)$ samples the risk field around the vehicle footprint (e.g., max- or area-average over a swept polygon). 
We use $(\alpha,\beta,\gamma)$ to trade off risk avoidance, control effort, and comfort, so no additional training is required.

\section{Conclusion}

This work introduced \textbf{MC-Risk}, a modular BEV risk field that couples learned multimodal forecasting with explainable analytic components to provide early, calibrated, and planner-ready risk maps. The field comprises a velocity/curvature-aware motorized-agent kernel, a forward-biased anisotropic VRU model, and a topology-informed road penalty term. Together, these yield a representation suitable for direct consumption by downstream planners. We performed a standardized, visibility-consistent evaluation on the RiskBench collision subset and achieved the strongest overall risk localization ability with the highest stability.

\paragraph{Limitations}
Closed-loop planning results using MC-Risk are not yet reported. Future directions include (i) closed-loop planning evaluations with on-policy setup in dense traffic; (ii) transfer to real-world datasets and hardware-in-the-loop tests.
{
    \small
    \bibliographystyle{IEEEtran}
    \bibliography{literature}
}

\appendix

\section{Experiments compute resources}
The experiments are conducted on a laptop with an Intel(R) Core(TM) i7-9750H CPU @ 2.60GHz with 13GB of system memory, which is used both for testing and debugging. No GPU resource is required for the implementation.

\section{Societal Impacts}

MC-Risk can advance road safety by delivering earlier, cleaner risk cues without sacrificing planner compatibility, especially for pedestrians and cyclists. Its forward-biased VRU modeling and topology-aware road layer help vehicles anticipate conflicts at crosswalks, merges, and opposing lanes, reducing high-severity collisions. Because the method is modular and light-weight, it is practical for embedded systems and accessible to most operating environments. Finally, a benchmarked risk map approach enables reproducible research and further comparison in this area.

\section{Evaluation metrics in detail}

We evaluate along two complementary axes: spatial accuracy and temporal consistency, following RiskBench \cite{kung2024riskbench} with adaptations from Pao et al.\ \cite{pao2024potential}.

\paragraph{Detection Accuracy.}
Let $TP$, $FP$, and $FN$ denote true positives, false positives, and false negatives identifying risk objects, respectively. Precision and recall are
\begin{equation}
\mathrm{precision}=\frac{TP}{TP+FP},\qquad
\mathrm{recall}=\frac{TP}{TP+FN}.
\end{equation}
The F1-score, the harmonic mean of precision and recall, is
\begin{equation}
F1=\frac{2\cdot \mathrm{precision}\cdot \mathrm{recall}}{\mathrm{precision}+\mathrm{recall}}.
\end{equation}

\paragraph{Overall Thresholded F1 (OT-F1).}
We sweep a decision threshold $\tau$ over all unique predicted risk scores and compute $F1(\tau)$. The overall thresholded score is
\begin{equation}
\mathrm{OT\mbox{-}F1}=\max_{\tau}\;F1(\tau).
\end{equation}

\paragraph{PIC.}
The PIC metric \cite{kung2024riskbench,pao2024potential} penalizes late errors more strongly. Using per-frame $F1_t$ (at the global threshold that maximizes OT-F1), we define
\begin{equation}
\label{eq:PIC}
\mathrm{PIC}=-\sum_{t=1}^{T}\exp\!\left(-\frac{T-t}{T}\right)\cdot \log\!\big(F1_t\big).
\end{equation}
Lower values indicate earlier and more consistent correct identifications.

\paragraph{Weighted Multi-Object Tracking Accuracy (wMOTA).}
To measure temporal stability with class imbalance, we adopt wMOTA from \cite{pao2024potential}. For each frame $t$,
\begin{equation}
PM_t=w_p\big(FN_t+IDsw_t^{p}\big),\qquad
NM_t=w_n\big(FP_t+IDsw_t^{n}\big),
\end{equation}
where $IDsw_t^{p}$ and $IDsw_t^{n}$ count identity switches among ground-truth risky and non-risky actors, respectively, and $w_p,w_n$ are class weights. With $GT_t^{p}$ and $GT_t^{n}$ the ground-truth counts,
\begin{equation}
\label{eq:wMOTA}
\mathrm{wMOTA}=1-\frac{\sum_t\big(PM_t+NM_t\big)}{\sum_t\big(w_p\,GT_t^{p}+w_n\,GT_t^{n}\big)}.
\end{equation}
Higher wMOTA reflects fewer misses and switches, accounting for the prevalence of non-risky actors.

\end{document}